\documentclass[conference]{IEEEtran}
\IEEEoverridecommandlockouts
\usepackage{cite}
\usepackage{amsmath,amssymb,amsfonts}
\usepackage{algorithmic}
\usepackage{graphicx}
\usepackage{textcomp}
\usepackage[colorlinks,citecolor=red]{hyperref}
\usepackage{multirow}
\usepackage{array}
\usepackage{subfigure}
\usepackage{booktabs}
\usepackage{amsmath}
\usepackage{algorithm}    
\usepackage{algorithmic}   
\usepackage{wrapfig}
\usepackage{hyperref} 
\usepackage{bm}
\usepackage{caption}

\def\BibTeX{{\rm B\kern-.05em{\sc i\kern-.025em b}\kern-.08em
    T\kern-.1667em\lower.7ex\hbox{E}\kern-.125emX}}

\def\aboveeq{\setlength{\abovedisplayskip}{6pt}}
\def\downeq{\setlength{\belowdisplayskip}{6pt}}

\begin{document}

\title{Optical Flow Based Real-time Moving Object Detection in Unconstrained Scenes\\
{\footnotesize \textsuperscript{*}}
\thanks{}
}

\author{

\IEEEauthorblockN{1\textsuperscript{st} Junjie Huang}
\IEEEauthorblockA{\textit{Institute of Automation} \\
\textit{Chinese Academy of Sciences}\\
Beijing, China \\
huangjunjie2016@ia.ac.cn}
\and
\IEEEauthorblockN{3\textsuperscript{rd} Wei Zou}
\IEEEauthorblockA{\textit{Institute of Automation} \\
\textit{Chinese Academy of Sciences}\\
Beijing, China \\
wei.zou@ia.ac.cn}
\and

\IEEEauthorblockN{3\textsuperscript{rd} Zheng Zhu}
\IEEEauthorblockA{\textit{Institute of Automation} \\
\textit{Chinese Academy of Sciences}\\
Beijing, China \\
zhuzheng2014@ia.ac.cn}
\and
\IEEEauthorblockN{4\textsuperscript{th} Jiagang Zhu}
\IEEEauthorblockA{\textit{Institute of Automation} \\
\textit{Chinese Academy of Sciences}\\
Beijing, China \\
zhujiagang2015@ia.ac.cn}

}

\maketitle

\begin{abstract}
Real-time moving object detection in unconstrained scenes is a difficult task due to dynamic background, changing foreground appearance and limited computational resource. In this paper, an optical flow based moving object detection framework is proposed to address this problem. We utilize homography matrixes to online construct a background model in the form of optical flow. When judging out moving foregrounds from scenes, a dual-mode judge mechanism is designed to heighten the system's adaptation to challenging situations. In experiment part, two evaluation metrics are redefined for more properly reflecting the performance of methods. We quantitatively and qualitatively validate the effectiveness and feasibility of our method with videos in various scene conditions. The experimental results show that our method adapts itself to different situations and outperforms the state-of-the-art methods, indicating the advantages of optical flow based methods.
\end{abstract}

\begin{IEEEkeywords}
component, formatting, style, styling, insert
\end{IEEEkeywords}

\section{Introduction}
In this paper, we study the detection of moving object. Aiming at detecting moving objects from complex scenes, many methods have been proposed and developed in depth. As moving object is defined according to its state of motion, it can not be commendably detected by a feature based well-trained classifier like \cite{b3}. This common task is handled by some frameworks, which can be classified roughly into two categories: one is analyzing foreground and background together to discriminate them into two classes \cite{b8}\cite{b18}. The other is to obtain a discriminant background model for judging out the foreground points. For example, Tom et.al, \cite{b5} used statistical models Dirichlet process Gaussian mixture model (DP-GMM). Cui et.al. \cite{b1} and Zhou et.al. \cite{b26} modeled the background as a low rank matrix. And the others used Fuzzy Models \cite{b9}, Robust Subspace Models \cite{b4}, Sparse Models \cite{b6}, Optical Flow Velocity Field Models \cite{b25}, et.al. Methods mentioned above, to some extent, can reach a certain level foreground extraction. However, they mainly work under some strong constraints like under stationary scenes \cite{b5}\cite{b15}\cite{b25}, using batch processing \cite{b1}\cite{b8}\cite{b18}\cite{b26}, or needing global optimization \cite{b1}\cite{b8}\cite{b26}. 

To get rid of these constraints, we propose an optical flow based framework. The framework adopts the background modeling method but models the background online as well as in the scenes simultaneously including background and foregrounds, which is different from \cite{b25}. We firstly estimate the optical flow field by performing algorithm FlowNet2.0 \cite{b7}. Then we estimate a intermediate variable(i.e. the homography matrix) which can give a parametric description of the sensor's motion. Unlike many other works \cite{b10}\cite{b11}\cite{b22}\cite{b23}, who estimate the homography matrix using point pairs obtained by point tracking algorithm LK \cite{b14} or KLT \cite{b19}, we obtain point pairs using the optical flow field directly. This can avoid introducing extra computation cost and avoid introducing unreliable information as the tracking algorithms LK and KLT are done without global optimization. Finally the background is modeled in the form of optical flow using the homography matrix. 

Subsequently, the moving foregrounds are judged out by setting a threshold for the difference between the optical flow provided by optical flow estimating algorithm and that provided by the background model. To increase the accuracy of judgment and strengthen the system's adaptation to different situations, a dual-mode judge mechanism is introduced in this work to deal with the problem caused by the sensor's evident zooming(the details are described in Section~\ref{sec:fe}). 

In experiment part, two evaluation metrics are redefined. Because if the F-Measure evaluation metric is defined as in \cite{b20} et.al, the results in the frames that contain small foreground add little impact to the video-level result. We calculate frame-level precision, recall and F-Measure first, and the video-level result is obtained by averaging over all frames in the same video. In this way, we enable the evaluation metrics to deal with some videos that contain unbalanced size of foreground in different frames, and to more properly reflect the methods' ability of detecting foreground. We test the robustness of the method using ten videos with various scene conditions. Our method qualitatively and quantitatively outperforms state-of-the-art algorithms in this test. Moreover, we also test the efficiency of the proposed framework in depth and offer some advice for practical applications.

The contributions of this work are as follows: Firstly, the proposed background modeling method is performed online and is efficient enough for real-time application. Meanwhile, the background model constructed by our method is more precise than that by the existing methods. Secondly, a dual-mode judge mechanism is introduced to strengthen the system's adaptation to different situations. Thirdly, we redefine two evaluation metrics to make them more convictive and demonstrate the effectivity of our method through comprehensive experiments.

The remainder of this paper is organized as follows. Section~\ref{sec:RW} reviews the related work. Our detection framework based on optical flow is detailedly introduced in Section~\ref{sec:MT} and its effectiveness is verified in Section~\ref{sec:EP} by comprehensive experiments. Finally, Section~\ref{sec:CC} is devoted to conclusions.

\section{Related Work}
\label{sec:RW}

In this section, we review recent algorithms for moving object detection in terms of several main modules:  Gaussian model based, optical flow model based and optical flow gradient based.

\textbf {Gaussian Model Based.} The method proposed in \cite{b22} used Dual-Mode Single Gaussian Model (SGM) to model the background in grid-level, and utilized homography matrixes between consecutive frames to accomplish motion compensation by mixing models. Foreground was figured out by estimating the feature's conformity to the corresponding SGM. Benefitting from Dual-Mode SGM, the method can reduce the foreground's pollution to the background models. Analogously, Yun and Jin \cite{b23}, and Kurnianggoro et.al, \cite{b11} used a foreground probability map and simple pixel-level background models respectively to fine-tune the result obtained in \cite{b22}. Method in \cite{b13} is based on SGM and interpolated a full covariance matrix of the pixel models to achieve the motion compensation. The background model constructed and updated by these methods lack a reflection to the essence of the problem .They are sensitive to parameters and lack of robustness to different scenes.

{\bf Optical Flow Model Based.} Kurnianggoro et.al, \cite{b10} modeled the background using zero optical flow vectors instead. After using a homography matrix to align the previous frame, dense optical flow was estimated between the result of aligning and the current frame. Finally a simple optical-flow magnitude threshold was used to judge out the foreground points. As the homography matrixes are only used for aligning, the background model and the judge mechanism constructed by this method are too simple to deal with intricate unconstrained scenes.

\begin{figure}[htpb]
\begin{center}
\includegraphics[width=6cm]{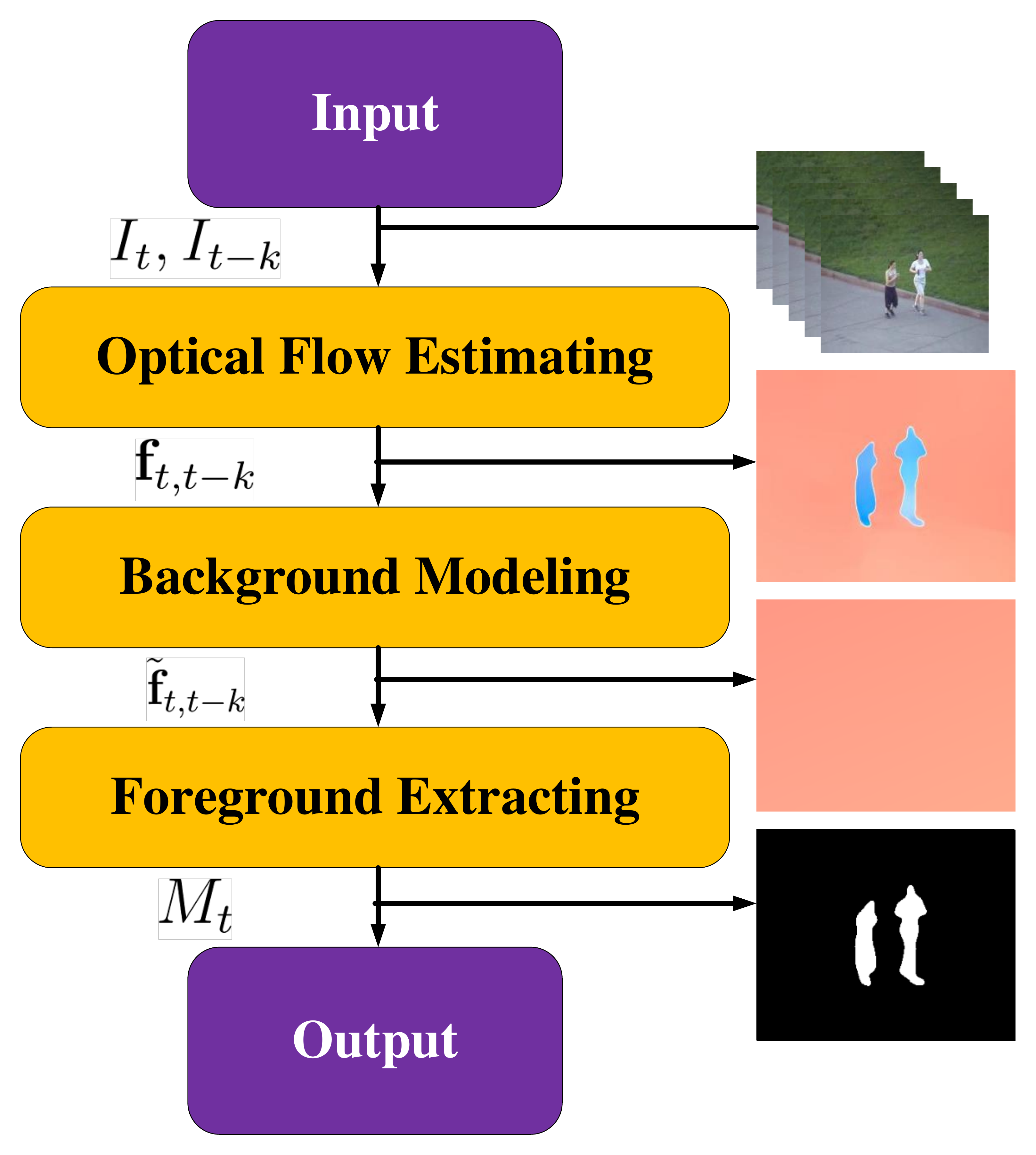}
\end{center}
\caption{Visualization of the moving object detection framework.}
\label{fig1}
\end{figure}

\textbf {Optical Flow Gradient Based.} There are some other methods that do not depend on any background models. They constructed the contour of foreground based on detecting large gradient points in dense optical flow field. For example, Li and Xu \cite{b12} performed mathematical morphology operations on the initial contours to obtain closed boundaries. After that the maximal contour area was selected as the area of the moving object. This simple framework can be performed easily but also limits the method to simple scenes. Papazoglou and Ferrari \cite{b17} combined the optical flow's gradient and direction to generate a better contour. Then, an efficient inside-outside maps algorithm was performed to initially figure out the foreground points, which was finally fine-tuned by global optimization. The short coming is that the inside-outside maps algorithm can obtain reasonable result only in simple scenes that contain a single object. Moreover, the optimization operation makes it inefficient.

\section{Methodology}
\label{sec:MT}
The framework of our online detection method for moving objects in dynamic scenes is shown in Fig.~\ref{fig1}. There are mainly three processes: optical flow estimating, background modeling and foreground extracting. In the following, each step of the framework is introduced in detail.

\subsection{Optical Flow Estimating}
\label{sec:OE}
Taking into account speed and accuracy, FlowNet2.0 \cite{b7} is used to estimate the optical flow vectors $\textbf f_{t,t-k}={\left[ \begin{array}{ccc}u&v&0\end{array} \right ]}^T$, which project 2D locations $\textbf p_t={\left[ \begin{array}{ccc}x&y&1\end{array} \right ]}^T$ in frame $t$ to the locations $\textbf p_{t-k}=\textbf p_t+\textbf f_{t,t-k}$ in specified frame $t-k$. The optical flow vectors are used as the main feature in the following procedures.

\subsection{Background Modeling}\label{CB}
In our framework, background model is constructed in the form of optical flow utilizing homography matrixes. To obtain the homography matrixes $H_{t\to t-k}$, we establish the equation \eqref{eq2} that reflects the same effectiveness of two conversion processes: transforming via homography matrixes and transforming via optical flow. 
{\aboveeq
\downeq
\begin{equation}
\label{eq2}
H_{t\to t-k}*\textbf{P}_t=\textbf{P}_t+\textbf {F}_{t, t-k}
\end{equation}}
where $\textbf{F}_{t,t-k}={\left[ \begin{array}{cccc}\textbf{f}_{t,t-k}^1&\textbf{f}_{t,t-k}^2&\dots&\textbf{f}_{t,t-k}^n\end{array} \right ]}$, $\textbf{P}_t={\left[ \begin{array}{cccc}\textbf{p}_t^1&\textbf{p}_t^2&\dots&\textbf{p}_t^n\end{array} \right ]}$. As a homography matrix contains $8$ free variables, $n=4$ different sample points are used.

The least square solution of equation \eqref{eq2} is solved and optimized by RANSAC \cite{b2} to obtain a more reliable result. We perform RANSAC with the sample point number $n=4$ and the iterations $iter=50$, which can provide an ideal success rate above $1-(1-(1/2)^4)^{50}=0.96$, given the assumption that the background occupies area more than half of the images. To improve the efficiency of RANSAC algorithm, the sampling points  are sparsely sampled in 2D image plane. Specifically, the images are partitioned into 16 pieces and $n$ of them are randomly selected, then one point is randomly chosen inside each selected pieces. Finally, the ideal background model is constructed in the form of optical flow that is calculated by the following equation: 
{\aboveeq
\downeq
\begin{equation}
\label{eq3}
\tilde{\textbf f}_{t,t-k}=H_{t\to t-k}*\textbf p_t-\textbf p_t
\end{equation}}
where $\tilde{\textbf f}_{t,t-k}={\left[ \begin{array}{ccc}\tilde{u}&\tilde{v}&0\end{array} \right ]}^T$ is the ideal optical flow vector of each background point.

\subsection{Foreground Extracting}\label{CB}
\label{sec:fe}
Subsequently, based on the background model, we judge out the foreground points by utilizing a dual-mode judge mechanism. Under normal conditions, we apply a adaptive threshold to the difference between the ideal background optical flow and the actual optical flow, and obtain a foreground mask as described in \eqref{eq4}:
{\aboveeq
\downeq
\begin{equation}
\label{eq4}
M_{t}=\{ Points\mid d_v>{T_a}\},\,\,d_v=||\textbf f_{t,t-k}-\tilde{\textbf f}_{t,t-k}||_2
\end{equation}}
where $d_v$ is the 2-norm of the complement vector. The adaptive threshold is defined as:
{\aboveeq
\downeq
\begin{equation}
\label{eq5}
T_a=a_1+a_2*\sqrt{{H_{t\to t-k}(1,3)}^2+{H_{t\to t-k}(2,3)}^2}
\end{equation}}
where $a_1$ and $a_2$ are the hyper-parameters used to control the magnitude of threshold. $a_1$ is the static component part corresponding to the destabilization caused by the sensor's resolution or the optical flow's precision. $a_2$ is used to introduce the dynamic component part, and we use a high threshold when the sensor moves fast. And the magnitude of the homography matrix elements ${H_{t\to t-k}(2,3)}$ and ${H_{t\to t-k}(2,3)}$ linearly reflects the speed of the sensor's motion.

\begin{figure*}[h]
\subfigure[The mixture optical flow field.] { \label{fig:3a} 
\begin{minipage}[tb]{0.30\textwidth}
\includegraphics[scale=0.31]{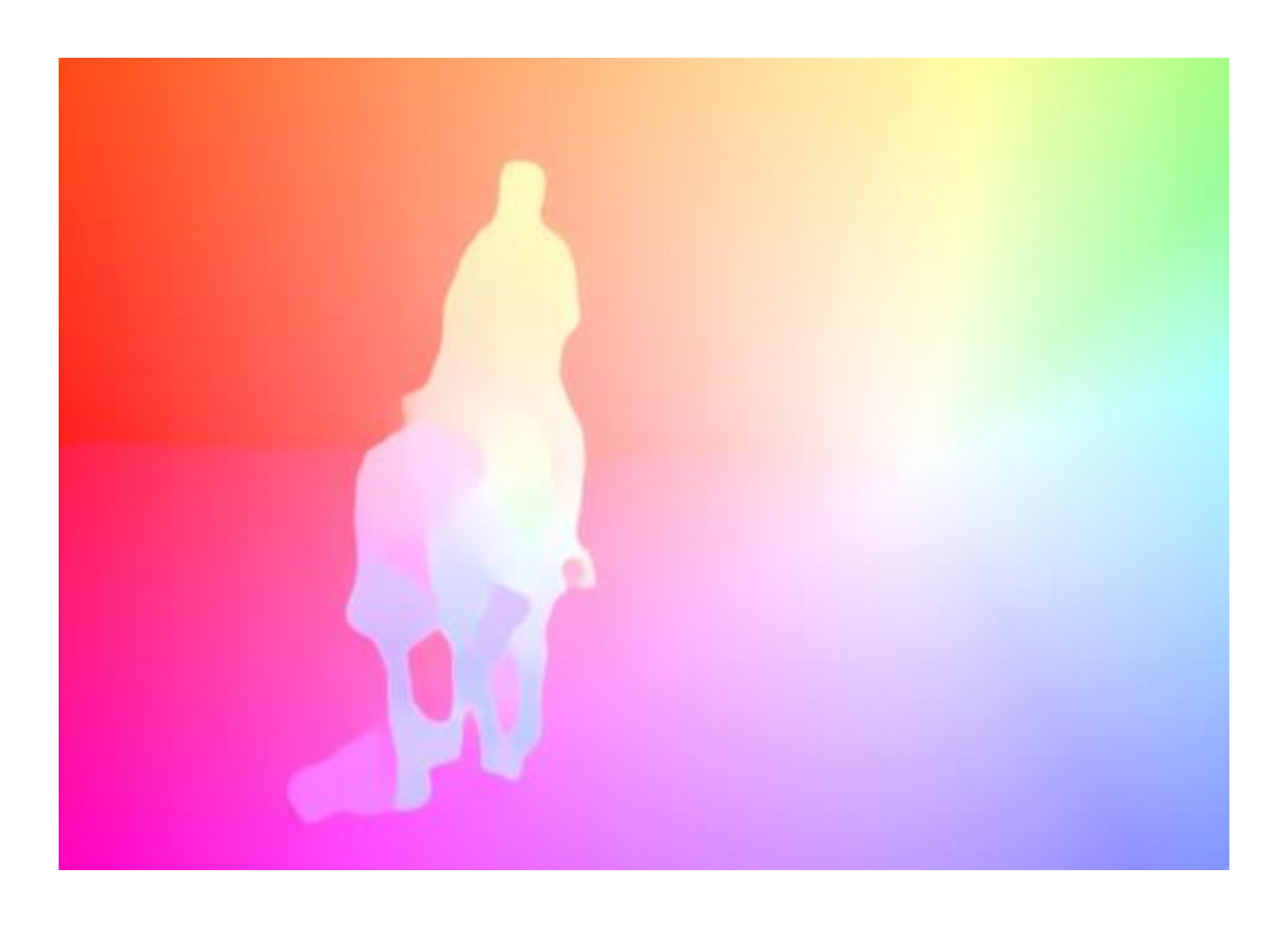}
\end{minipage}
}
\subfigure[The intensity map of $d_v$.] { \label{fig:3b} 
\begin{minipage}[tb]{0.34\textwidth}
\includegraphics[scale=0.31]{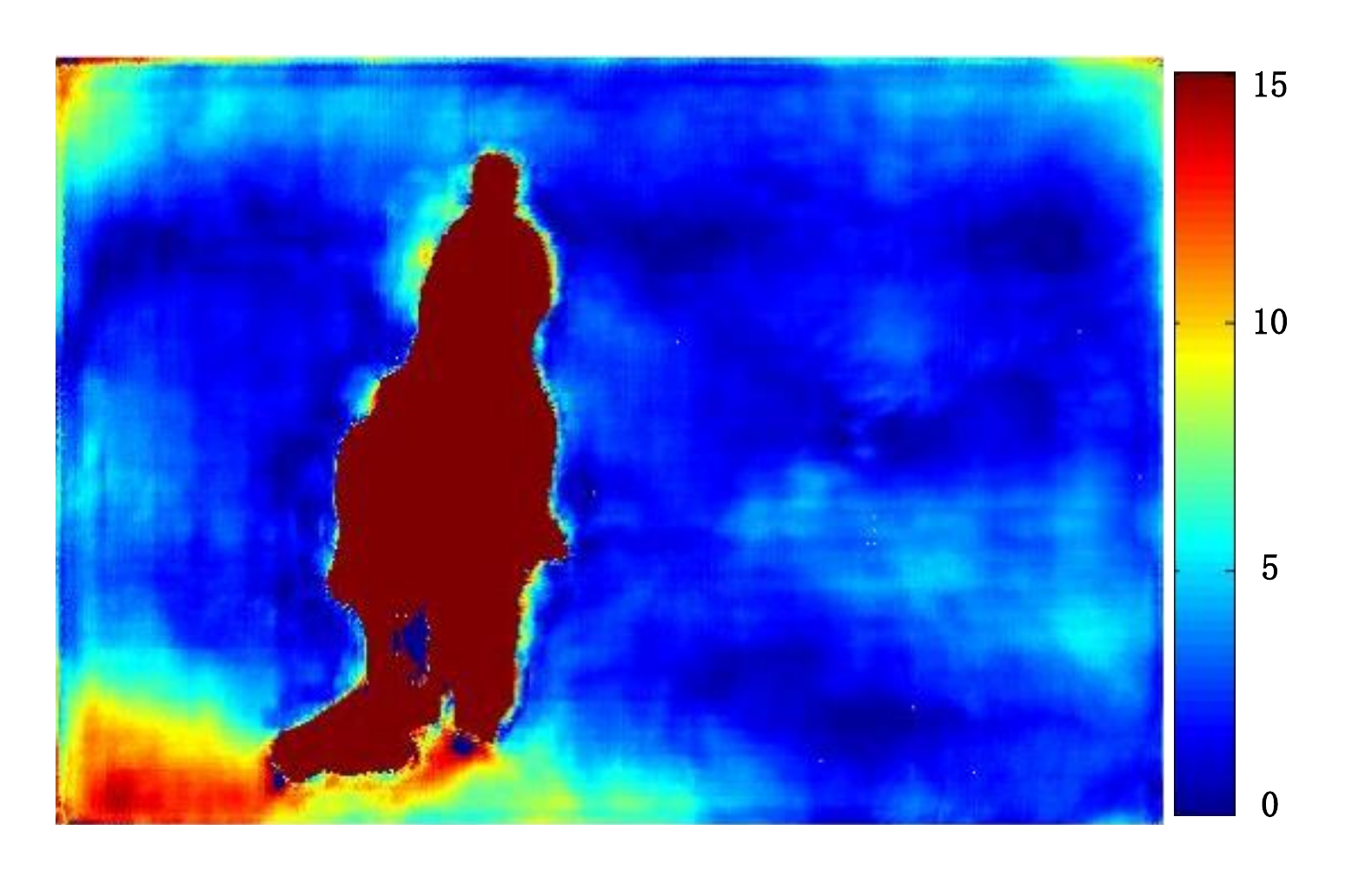}
\end{minipage}
}
\subfigure[The result obtained by \eqref{eq4}.] { \label{fig:3c} 
\begin{minipage}[tb]{0.30\textwidth}
\includegraphics[scale=0.31]{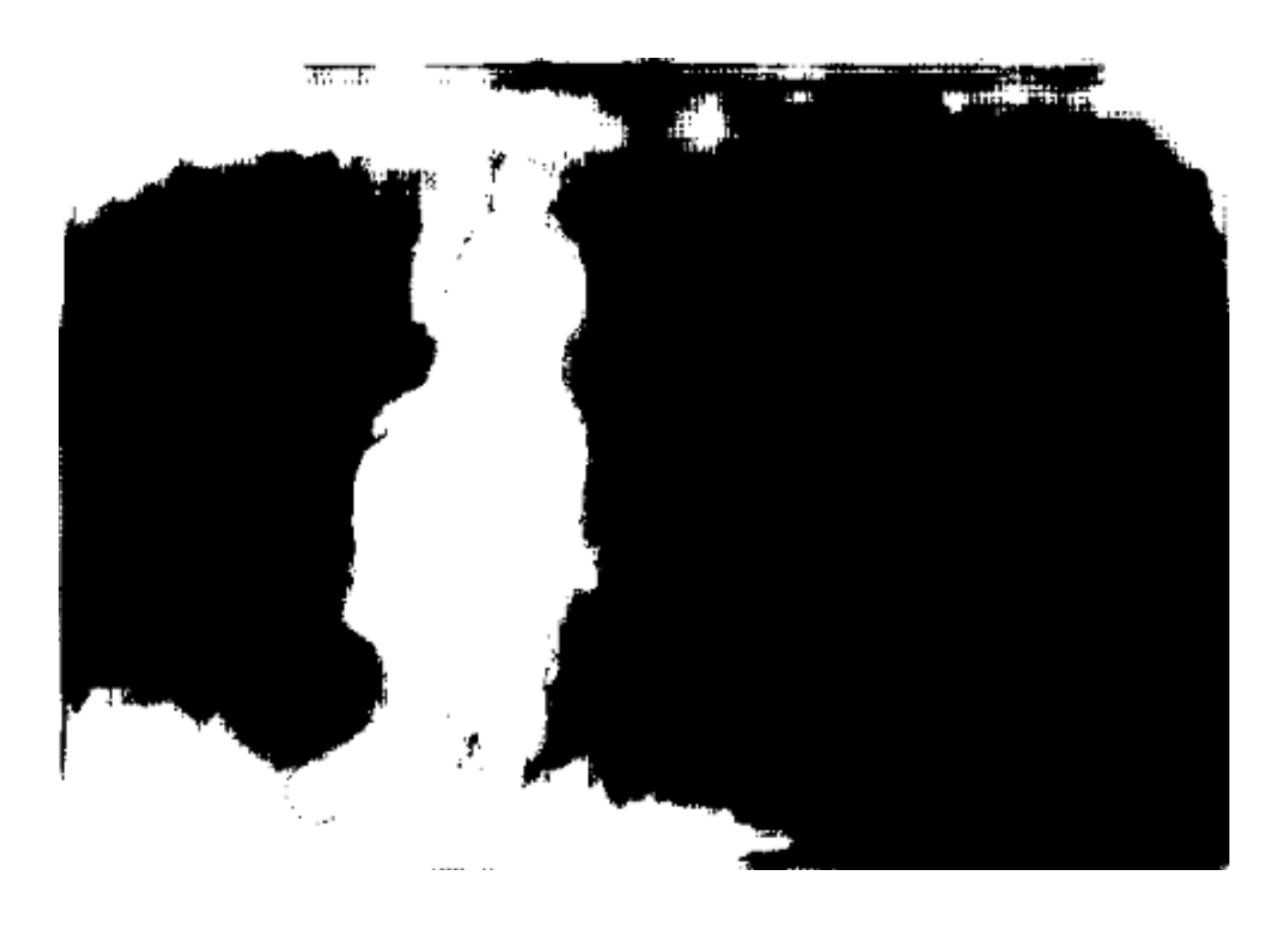}
\end{minipage}
}

\centering\subfigure[The ideal background optical flow field.] { \label{fig:3d} 
\begin{minipage}[bt]{0.30\textwidth}
\includegraphics[scale=0.31]{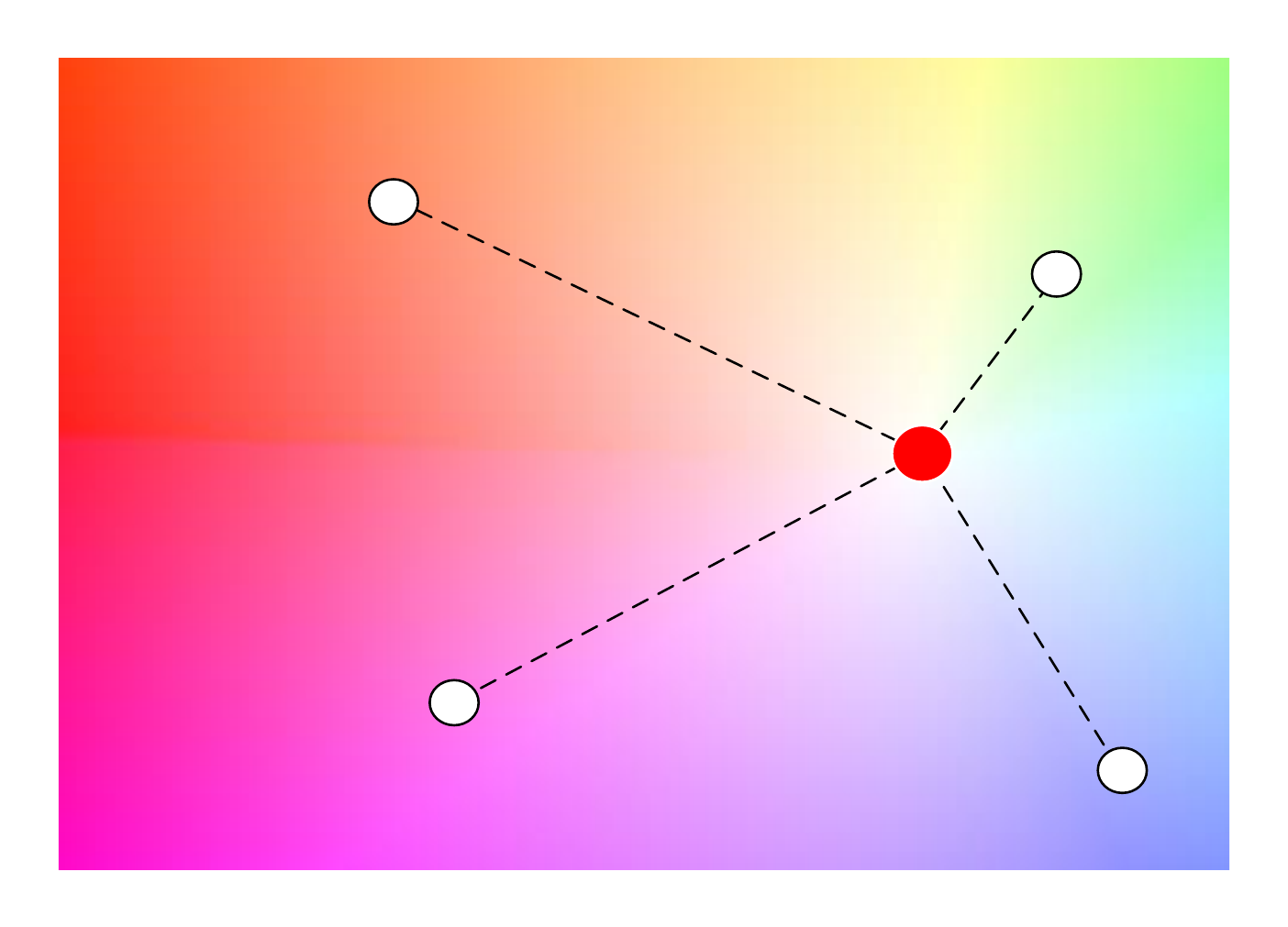}
\end{minipage}
}
\centering\subfigure[The intensity map of $d_c$.] { \label{fig:3e} 
\begin{minipage}[bt]{0.34\textwidth}
\includegraphics[scale=0.31]{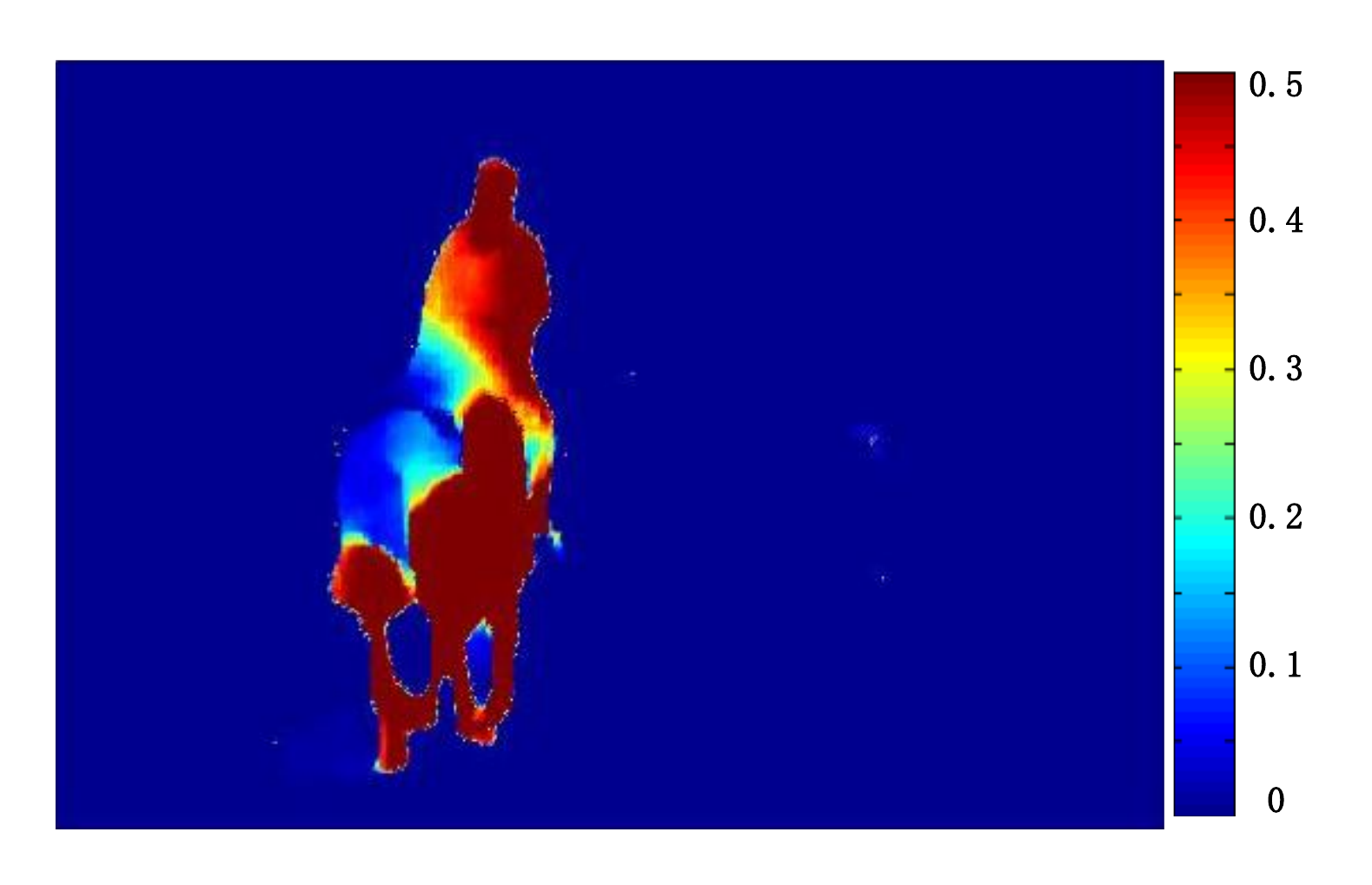}
\end{minipage}
}
\centering\subfigure[The result obtained by \eqref{eq8}.] { \label{fig:3f} 
\begin{minipage}[bt]{0.30\textwidth}
\includegraphics[scale=0.31]{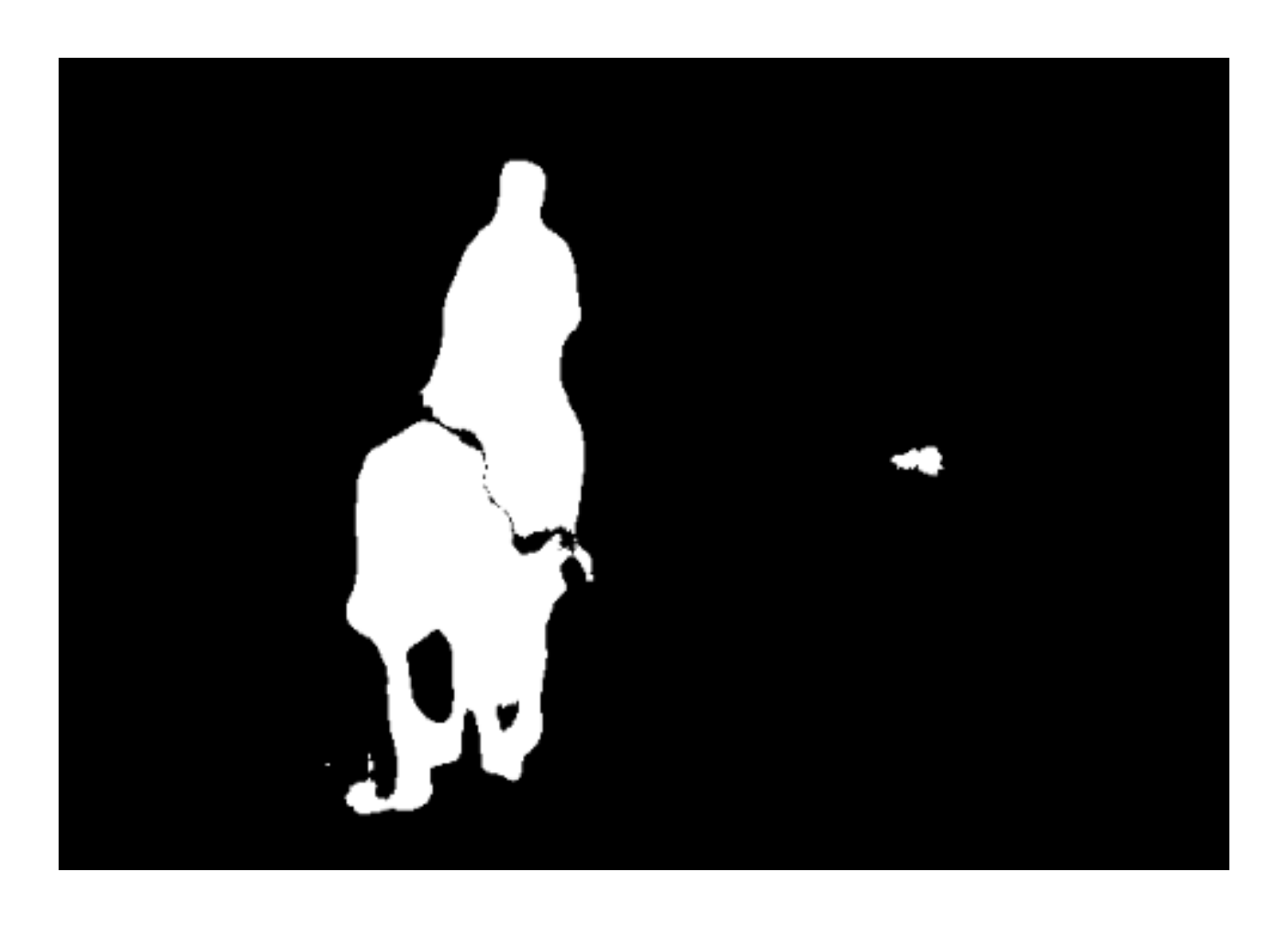}
\end{minipage}
}
\caption{Illustration for the evident zooming situation. In Fig.~\ref{fig:3d}, the white points denote the sampling background points and the red one denotes the intersection point.} 
\label{fig3} 
\end{figure*}

It is reasonable to obtain moving foregrounds in this way under most situations except that there is evident zooming composition in the scene change. Under this special situation, the spacial distribution of the background optical flow is unbalanced in amplitude, which will cause unbalanced spacial distribution of the difference between ideal optical flow and true optical flow, just as shown in Fig.~\ref{fig:3b}. Thus a fixed threshold calculated by the aforementioned method is not effective for judging out the foreground properly. 

We firstly detect this situation by analyzing the direction and amplitude of the background optical flow. As shown in Fig.~\ref{fig:3d}, under the evident zooming situation, the directions of background points' optical flow will intersect at the same vanishing point $\textbf p_0$, which is inside the image and the variation of the optical flow amplitudes in background area will exceed a certain threshold. In this work, we utilize the $n$ sample points which are used to calculate the homography matrix on behalf of the background points. With these background points, coordinates of the vanishing point $textbf p_0$ is calculated by geometrical analysis and least square regression:
{\aboveeq
\downeq
\begin{equation}
\label{eq6}
\textbf p_0=(A^T*A)^{-1}*A^T*B
\end{equation}}
where 
\begin{align*}
&A={\left[ \begin{array}{cc}V& -U\end{array} \right ]}\\
&U={\left[ \begin{array}{cccc}u^1& u^2&\dots&u^n\end{array} \right ]}^T\\
&V={\left[ \begin{array}{cccc}v^1& v^2&\dots&v^n\end{array} \right ]}^T\\
&B={\left[ \begin{array}{c}V\odot X-U\odot Y\end{array}\right ]}\\
&X={\left[ \begin{array}{cccc}x^1& x^2&\dots&x^n\end{array} \right ]}^T\\
&Y={\left[ \begin{array}{cccc}y^1& y^2&\dots&y^n\end{array} \right ]}^T 
\end{align*}
$\odot$ denotes the matrix multiplication of elements.
Then a judgment indicator is defined as:
{\aboveeq
\downeq
\begin{equation}
\label{eq7}
idx=(\textbf p_0\in \mathbb{W}*\mathbb{H}) \&\& (||\nabla ||\textbf f_{t,t-k}||_2 ||_2>T_g )
\end{equation}}
where $\mathbb{W}=[0,w]$, $\mathbb{H}=[0,h]$ denotes the ranges of coordinates inside image. For $idx$, a value of 1 indicates that there is a evident zooming  composition, on the other hand, a value of 0 indicates there isn't.

The ensuing question is how to properly judge out the foreground points when the magnitude threshold loses efficacy. Because the direction of true optical flow is highly identical to that of ideal optical flow in background area as shown in Fig.~\ref{fig:3e}, the foreground is extracted by a different judge mechanism:
{\aboveeq
\downeq
\begin{equation}
  \label{eq8}
  M_{t}=\{ Points\mid d_c<{T_c}\},\,\,d_c=cos(\textbf f_{t,t-k},\tilde{\textbf f}_{t,t-k})
\end{equation}}
In Equation \eqref{eq8}, the cosine value of the angle between the true optical flow $\textbf f_{t,t-k}$ and the ideal background optical flow $\tilde{\textbf f}_{t,t-k}$ is calculated, and is applied to judge out the foreground points by comparing it with a threshold $T_c$. Just as shown in Fig.~\ref{fig:3c} and Fig.~\ref{fig:3f}, judging according to $d_c$ is more efficient in situations with evident zooming than that according to $d_v$.

\begin{algorithm}[ht]
	\caption{Moving Object Detection}
	\begin{algorithmic}[1]
	\STATE\textbf{Input:} images $I_t$ and $I_{t-k}$
	\STATE estimating optical flow $\textbf f_{t,t-k}$ utilizing $I_t$ and $I_{t-k}$;
	\STATE establishing the equation \eqref{eq2} and solving it to obtain a homography matrix $H_{t\to t-k}$;
	\STATE obtaining the background model $\tilde{\textbf f}_{t,t-k}$ utilizing \eqref{eq3};
	\STATE judging out the evident zooming situation by \eqref{eq7};
	\IF{$idx=0$}
	\STATE extracting foreground mask $M_t$ utilizing \eqref{eq4}
	\ELSE
	\STATE extracting foreground mask $M_t$ utilizing \eqref{eq8}
	\ENDIF
	\STATE\textbf{Output:} foreground mask $M_t$
    \end{algorithmic}  
\end{algorithm}  
\section{Experiment}
\label{sec:EP}
The proposed method is implemented using Matlab, and is roundly tested with ten video sequences captured by unconstrained cameras: Playground1(PG1), Playground2(PG2), Skating1(SK1), Skating2(SK2), Walking(WK), Car1, Car2, Horse, Train and Highway(HW). Detecting moving object in these sequences is challenging, due to camera movement, irregular object movement, variational object appearance, bad weather and many other reasons. PG1, PG2, SK1 and WK sequence are from \cite{b24}, SK2, Train and HW are from \cite{b20}, Car2 and Horse are from \cite{b16}, Car1 is from \cite{b21} and annotated by our.

\begin{figure*}[htbp]
	\centering
	\subfigure[input image] { \label{fig:a} 
	\begin{minipage}[tb]{0.15\textwidth}
		\includegraphics[scale=0.157]{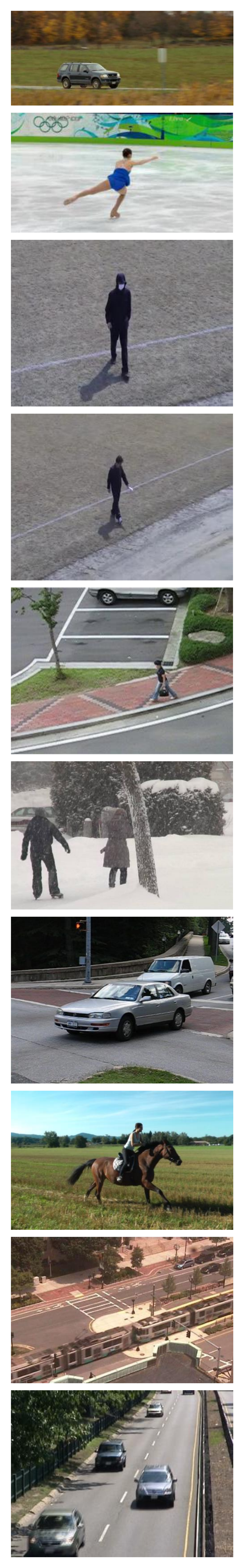}
	\end{minipage}}
	\subfigure[ground truth] { \label{fig:b} 
	\begin{minipage}[tb]{0.15\textwidth}
		\includegraphics[scale=0.157]{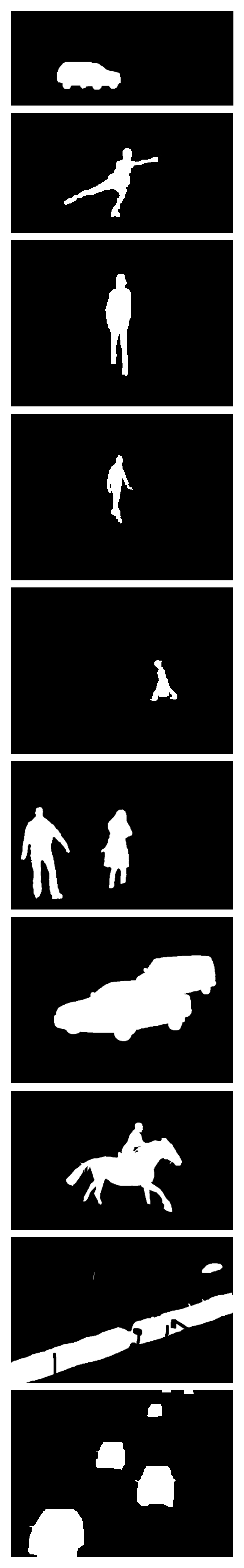}
	\end{minipage}}
	\subfigure[MCD5.8ms] { \label{fig:c} 
	\begin{minipage}[tb]{0.15\textwidth}
		\includegraphics[scale=0.157]{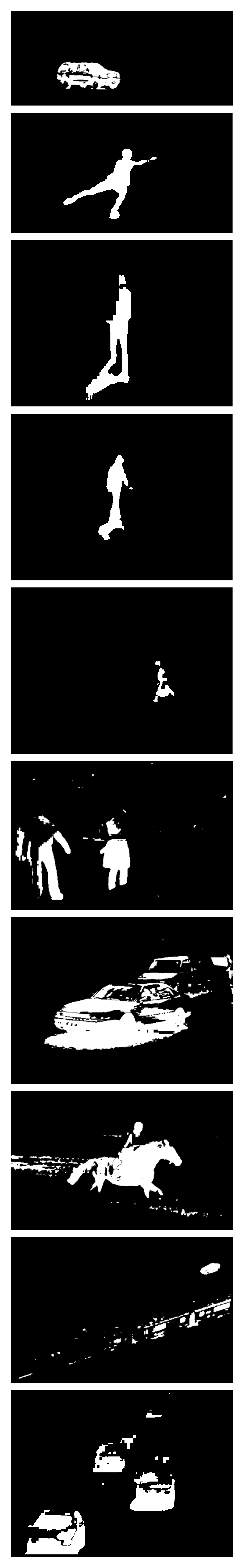}
	\end{minipage}}
	\subfigure[SA] { \label{fig:d} 
	\begin{minipage}[bt]{0.15\textwidth}
		\includegraphics[scale=0.157]{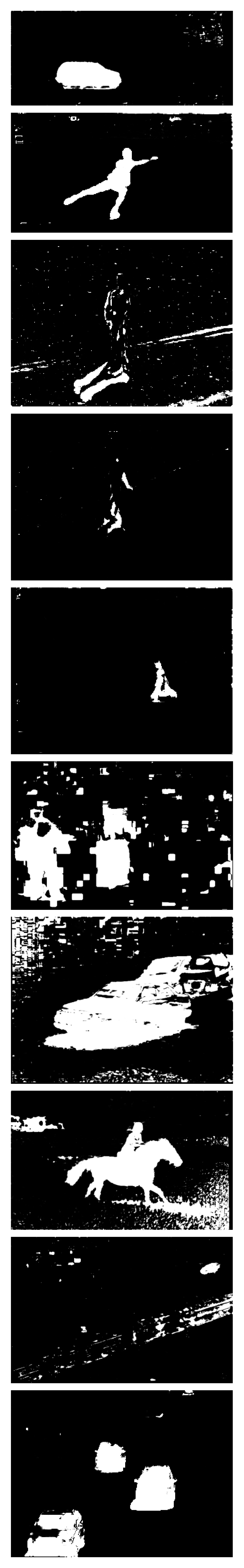}
	\end{minipage}}
	\subfigure[SCBU] { \label{fig:e} 
	\begin{minipage}[tb]{0.15\textwidth}
		\includegraphics[scale=0.157]{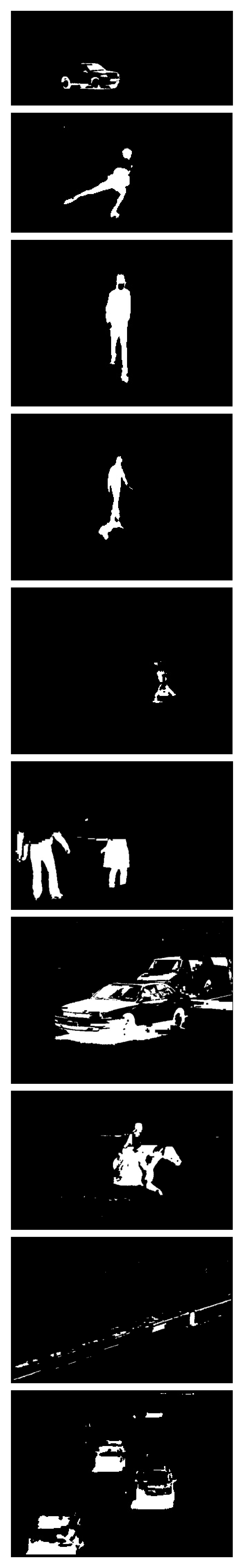}
	\end{minipage}}
	\subfigure[Ours] { \label{fig:b} 
	\begin{minipage}[tb]{0.15\textwidth}
		\includegraphics[scale=0.157]{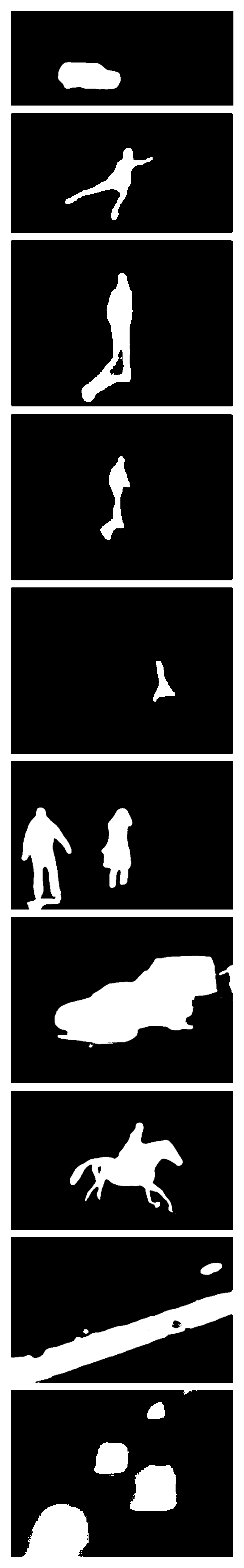}
	\end{minipage}}
	\vspace{1em}
	\caption{Qualitative results on some key frames from different videos. From top to bottom: Car1, SK1, PG1, PG2, WK, SK2, Car2, Horse, Train and HW. The first column shows input images, and the other columns show the results of the compared methods: (a) input image, (b) ground truth, (c) MCD5.8ms \cite{b22} , (d) Stochastic Approx(SA) \cite{b13} , (e) SCBU \cite{b24} and (f) Our method.} 
	\label{fig4}
\end{figure*}

\subsection{Evaluation Metrics}\label{AA}
In this subsection, we redefine two metrics for the method performance evaluation: F-Measure(FM), Success Rate(SA). Given true positive(TP), false positive(FP), false negative(FN) and  true negative(TN), F-Measure is defined as the harmonic mean of Precision(Pr) and Recall(Re) by:
{\aboveeq
\downeq
\begin{equation}
	FM= \frac{1}{N}\sum_{i=1}^NFM_i= \frac{1}{N}\sum_{i=1}^N \frac{2\times Pr_i\times Re_i}{Pr_i+Re_i}
\end{equation}}
where $Pr_i=\frac{TP_i}{TP_i+FP_i}$, $Re_i=\frac{TP_i}{TP_i+FN_i}$, $i$ denotes the sequence number of a frame in a video containing $N$ frames. F-Measure ranges from 0 to 1, where a value of 1 indicates that the prediction totally agrees with its ground truth, on the other hand, a value of 0 indicates total disagreement. 

Success Rate is used to more intuitively observe how well the method detects foreground. Given all $FM_i$ of a video sequence that contains $N$ images and a F-Measure threshold($T_{FM}$), Success Rate(SR) is defined as:
{\aboveeq
\downeq
\begin{equation}
	SR=\frac{1}{N}\sum_{i=1}^N(FM_i>T_{FM})
\end{equation}}
where $T_{FM}$ ranges from 0 to 1, and after the value of $T_{FM}$ changes, a curve is constructed as shown in Fig.~\ref{fig5}. The methods whose success rate curve are close to the top and right of the plot respectively have higher detecting success rate and higher detecting quality.

\subsection{Parameters setting}\label{AA}
Optical flow is estimated between frame $t$ and frame $t-5$, which means that the interval was set as $k=5$. For RANSAC algorithm, sampling point number $n$ and iterations $iter$ are set as in Section~\ref{CB} . For foreground judging, parameters are set as following: $T_g=0.032$, $a_1=0.1*k$, $a_2=0.3$ and $T_c=0.99$.

\subsection{Qualitative comparisons}\label{AA}
Our method is compared with the following state-of-the-art methods for moving object detection under an unconstrained camera: MCD5.8ms \cite{b22}, SA \cite{b13} and SCBU \cite{b24}. Fig.~\ref{fig4} shows the qualitative results on some key frames from the experimental video sequences.

The qualitatively comparative results can intuitively show the proposed method's adaptability to different challenges comparing with the other methods. As shown in SK2 and Train sequences, SGM base methods MCD5.8ms and SCBU perform poorly when the foreground color is slightly similar to the background. According to PG1, PG2 and SK2 sequences, SA can not deal with the challenges of slow motion and dynamic background. Benefitting from the optical flow based model and the dual-mode judge mechanism, the proposed method can export foreground with higher quality in all these scenes. According to the results of PG1, PG2, Car2 and HW sequences, our method is sensitive to shadow, which leads to some false positive results and has negative influence on the quantitative results.

\begin{table*}[htbp]
\begin{center}
\begin{tabular}{p{18mm}p{7.8mm}<{\centering}p{7.8mm}<{\centering}p{7.8mm}<{\centering}p{7.8mm}<{\centering}p{7.8mm}<{\centering}p{7.8mm}<{\centering}p{7.8mm}<{\centering}p{7.8mm}<{\centering}p{7.8mm}<{\centering}p{7.8mm}<{\centering}p{7.8mm}<{\centering}}
\hline
\specialrule{0em}{1pt}{2pt}
\textbf{Method}&\textbf{PG1}&\textbf{PG2}&\textbf{SK1}&\textbf{WK}&\textbf{SK2}&\textbf{Car1}&\textbf{Car2}&\textbf{Horse}&\textbf{Train}&\textbf{HW}&\textbf{AVG}\\
\hline
\specialrule{0em}{2pt}{2pt}
MCD5.8ms&0.356&0.546&\textbf{0.821}&0.729&0.595&0.618&0.260&0.672&0.260&0.433&0.529\\
\specialrule{0em}{1pt}{2pt}
SA&0.123&0.276&0.384&\textbf{0.774}&0.475&0.606&0.405&0.788&0.257&0.668&0.476\\
\specialrule{0em}{1pt}{2pt}
SCBU&0.543&\textbf{0.679}&0.759&0.672&0.632&0.356&0.120&0.633&0.127&0.459&0.506\\
\specialrule{0em}{1pt}{2pt}
Ours&\textbf{0.550}&0.653&0.643&0.677&\textbf{0.827}&\textbf{0.887}&\textbf{0.733}&\textbf{0.909}&\textbf{0.867}&\textbf{0.724}&\textbf{0.747}\\
\specialrule{0em}{1pt}{2pt}
\hline
\end{tabular}
\end{center}
\caption{The pixel-wise F-measure results for all of ten videos. AVG denotes Average.}
\label{tab2}
\end{table*}

\subsection{Quantitative comparisons}\label{AA}
We also quantitatively compare our method with the state-of-the-art methods: MCD5.8ms \cite{b22}, SA \cite{b13} and SCBU \cite{b24}. As shown in Table.~\ref{tab2}, the pixel-wise F-Measure of each video sequence is calculated and an average between videos is given in the back. While the existing methods perform worse in some situation, the proposed method outputs steadily when facing all different challenges. It is noteworthy that in PG1 video sequence, the pixel-wise F-Measure results of all method are rather low. We analyze the masks outputted by each method and find out that PG1 contains very tiny object with rather slow motion in the first $300$ frames, which is rather difficult for the systems to detect. Comparing our result with that in \cite{b24}, the redefined evaluation metrics can more properly reflect the performance of the methods.

\begin{figure}[htbp]
\centering
\includegraphics[width=6.2cm]{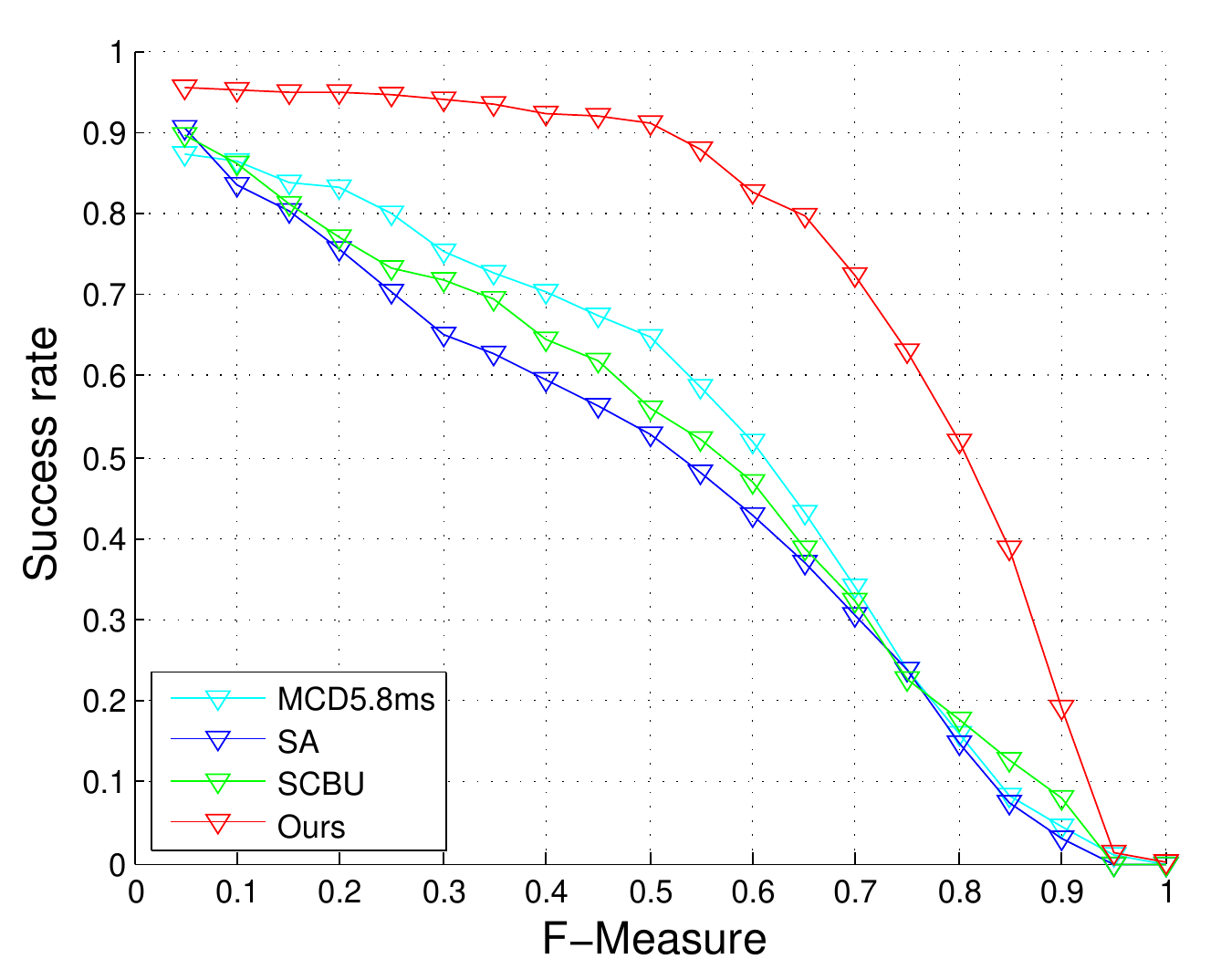}
\caption{Relation curves of F-Measure threshold and Success rate.}
\label{fig5}
\end{figure}

Fig.~\ref{fig5} illustrates the average success rate plots of these compared methods. The proposed optical flow based method outperforms other methods both in success rate and in quality. Given a specific threshold value of $FM=0.5$, the success rate of the proposed method is $0.92$, which is high enough for practical applications.

\subsection{Efficiency}\label{AA}

We measured the computation time per frame to evaluate the efficiency of the proposed method. Table.~\ref{tab5} shows the computation time  measured by Matlab on an Intel Core i5-7400 3.0GHz PC with video at a resolution of $320*240$. As shown in the result, the optical flow estimation process occupies eight out of ten total time consumption, and the foreground extracting process spent relatively less time. 

\begin{table}[htbp]
\begin{center}
\begin{tabular}{p{15mm}p{10mm}<{\centering}p{10mm}<{\centering}p{10mm}<{\centering}}
\hline
\specialrule{0em}{1pt}{2pt}
\textbf{Process} & \textbf{OE}&\textbf{MD}&\textbf{total}\\
\hline
\specialrule{0em}{1pt}{2pt}
\textbf{Time(ms)} &${123}^\star$&16 &139 \\
\hline
\end{tabular}
\end{center}
\caption{Time consumption of the proposed method. The entries show the time consumption of each process in the form of ms per frame. OE denotes Optical Flow Estimation, MD denotes Moving Object Detection, $\star$result is quoted from \cite{b7}.}
\label{tab5}
\end{table}

\begin{figure}[htbp]
\centering
\includegraphics[width=6.1cm]{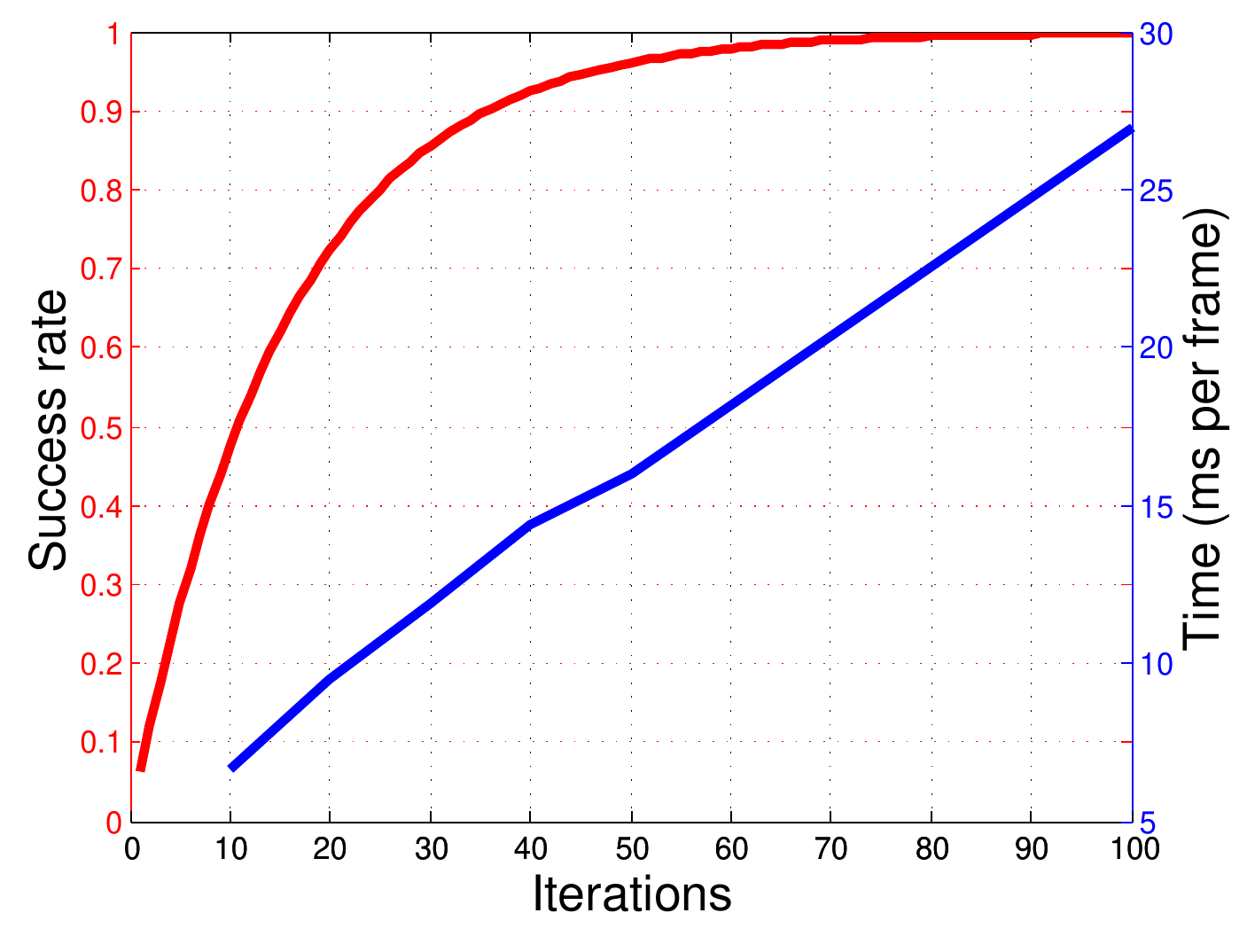}
\caption{Illustration of RANSAC algorithm iterations' effect. The red curve shows the success rate related to iterations and the blue curve shows the time consumption.}
\label{fig6}
\end{figure}

Further experiment shows that the time consumption in foreground extraction process has a linear correlation with the iterations of RANSAC algorithm. Fig.~\ref{fig6} illustrates the relation curve of iterations and time consumption as well as the relation curve of iterations and the ideal success rate of finding out the correct background model. When the iterations increase to $40$, the success rate has been above 0.9 and the speed of increasement has been slow obviously. So it is reasonable to set the iterations around $40$.

\section{Conclusion}
\label{sec:CC}
We propose an optical flow based framework for real-time moving object detection in unconstrained scenes. The background model is constructed in the form of optical flow utilizing homography matrixes, and a dual-mode judge mechanism is introduced to heighten the system's adaptation to different situations. In experiment part, two evaluation metrics are redefined for more properly reflecting the performance of the methods. The quantitative and qualitative results obtained by our framework outperform the state-of-the-art methods indicating the advantages of optical flow based method. Finally, the precision and frame rate of the optical flow estimation algorithm are the prerequisite of the success of our frame. With the development of optical flow estimation algorithm, the performance of our framework will correspondingly improve.

\end{document}